\documentclass[11pt]{article}

\PassOptionsToPackage{table}{xcolor}
\usepackage{acl}

\usepackage{times}
\usepackage{latexsym}
\usepackage[T1]{fontenc}
\usepackage[utf8]{inputenc}
\usepackage{microtype}
\usepackage{inconsolata}
\usepackage{graphicx}
\usepackage{booktabs}
\usepackage{multirow}
\usepackage{adjustbox}
\usepackage{pifont}
\usepackage{amsmath}
\usepackage{amssymb}
\usepackage{mathtools}

\DeclareMathOperator{\Hallucinate}{Hallucinate}
\newcommand{\correctmark}{\textcolor{green!70!black}{\ding{52}}}
\newcommand{\errormark}{\textcolor{red}{\ding{55}}}
\title{OmniHallu: Unified Hallucination Detection for Cross-Modal \\ Comprehension and Generation in Multimodal Large Language Models}

\author{
  Jianjiang Yang\textsuperscript{1} \quad
  Peihang Li\textsuperscript{2} \quad
  Shanqing Xu\textsuperscript{3} \quad Mengchen Qian\textsuperscript{3}\\
  \bfseries \quad
  Lu Zhang\textsuperscript{4} \quad
  Meng Luo\textsuperscript{5}\footnotemark[1] \\
  \normalfont
  \textsuperscript{1}The University of Manchester \quad
\textsuperscript{2}The University of Hong Kong\\
\textsuperscript{3}Huazhong University of Science and Technology\\
\textsuperscript{4}Shanghai Academy of Educational Sciences\\
\textsuperscript{5}National University of Singapore
}

\hypersetup{
  pdftitle={OmniHallu: Unified Hallucination Detection for Cross-Modal Comprehension and Generation in Multimodal Large Language Models},
  pdfauthor={Jianjiang Yang, Peihang Li, Shanqing Xu, Mengchen Qian, Lu Zhang, Meng Luo}
}

\begin{document}
\maketitle
\begingroup
\renewcommand{\thefootnote}{\fnsymbol{footnote}}
\footnotetext[1]{\raggedright Corresponding authors: Meng Luo (\texttt{mluo@u.nus.edu}).}
\endgroup
\setcounter{footnote}{0}

\begin{abstract}
While Multimodal Large Language Models (MLLMs) have achieved remarkable progress across diverse tasks, they suffer from hallucinations where generated outputs contradict or misrepresent input semantics. Existing research typically addresses hallucination detection within a single modality or task type, limiting generalizability. We introduce \textbf{OmniHallu}, a unified hallucination detection framework spanning both comprehension and generation tasks across image, video, and audio modalities. We contribute \textbf{OmniHallu-Bench}, a 10,000-sample benchmark with claim-level human annotations covering six cross-modal tasks: image-to-text (I2T), video-to-text (V2T), audio-to-text (A2T), text-to-image (T2I), text-to-video (T2V), and text-to-audio (T2A). Our multi-agent architecture decomposes model outputs into atomic claims, verifies them through modality-specific experts, and aggregates evidence via structured reasoning. We further propose a preference-optimized trainable verifier that approximates the multi-agent decision boundary, reducing expert calls by 66\% with minimal performance loss. Extensive experiments reveal a consistent modality-dependent performance gradient and provide fine-grained insights into cross-modal hallucination patterns.
\end{abstract}

\section{Introduction}
\label{sec:intro}

\begin{figure}[!t]
\centering
\includegraphics[width=0.98\columnwidth]{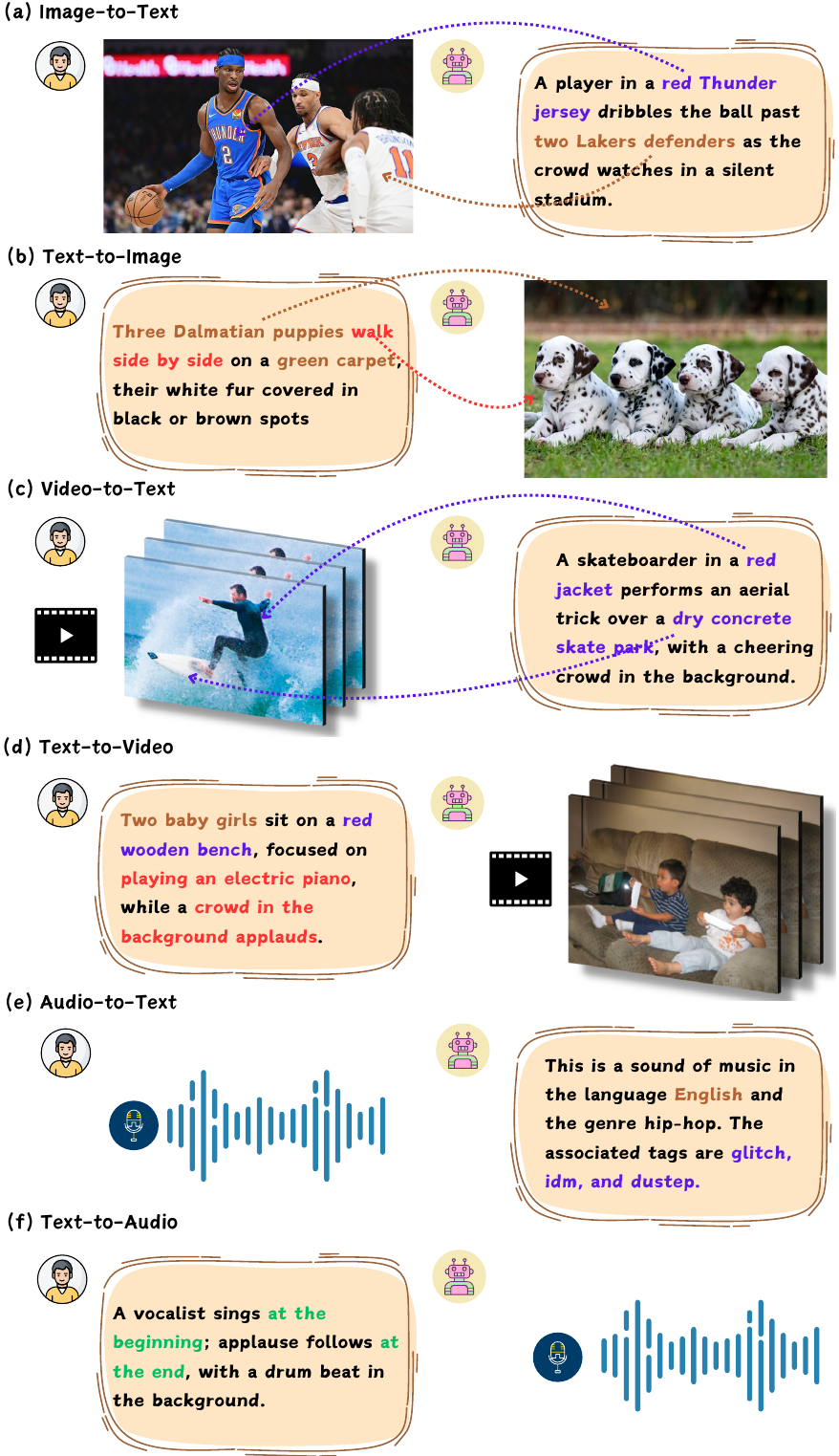}
\caption{MLLMs produce hallucinations in both comprehension and generation across modalities, spanning object, attribute, relation, and event hallucination types.}
\label{fig:intro}
\end{figure}

Multimodal Large Language Models (MLLMs) \citep{li2024survey,chen2025janusprounifiedmultimodalunderstanding,li2025mist,li2026unim,luo2024panosent} have achieved remarkable progress across vision, audio, and language tasks. However, these models frequently hallucinate: generating outputs that contradict or misrepresent the input \citep{bai2024hallucination,huang2024survey,lin2025fact,luo2026unveiling,luo2026dr}. Hallucinations pose a fundamental barrier to deploying MLLMs in safety-critical applications.

Existing hallucination detection methods predominantly target a single modality or task type. Image-focused benchmarks \citep{li2023evaluating,wang2024amberllmfreemultidimensionalbenchmark} do not cover video or audio; video-focused \citep{liu2023modelshallucinationsevaluatingfactuality,wang2024videohallucer} and audio-focused \citep{nishimura2024audio} evaluations similarly operate in isolation. Moreover, most work addresses only comprehension while neglecting generation tasks, despite both sharing common hallucination patterns rooted in insufficient perception and reasoning.

We study a unified claim-level detection protocol that enables side-by-side comparison across modalities and task directions, while also revealing which components transfer and where modality-specific verification remains necessary.

To this end, we introduce \textbf{OmniHallu}, a unified hallucination detection framework, and \textbf{OmniHallu-Bench}, a benchmark of 10,000 human-verified samples. Our method adapts the established decompose--verify--aggregate paradigm \citep{chen2024unifiedhallucinationdetectionmultimodal} to six bidirectional tasks: video verification emphasizes temporal and causal evidence, whereas audio verification relies on acoustic cues and a less mature tool ecosystem. Our contributions are:

\begin{itemize}
    \item \textbf{OmniHallu-Bench}: A 10,000-sample benchmark with claim-level human annotations spanning six cross-modal tasks (I2T, V2T, A2T, T2I, T2V, T2A) across four modalities.
    \item \textbf{Cross-modal systematization}: A modality-aware implementation of claim decomposition, specialized verification, and evidence aggregation, together with controlled analyses of component contributions, performance variations, and failure modes.
    \item \textbf{Preference-optimized verifier}: A compact trainable verifier aligned via GRPO that reduces expensive expert calls by 66\% with minimal performance loss.
\end{itemize}

\begin{table*}[t]
	\centering
    \caption{Comparison of hallucination benchmarks. ``Function'' indicates whether the benchmark supports fact-\textbf{C}hecking or hallucination \textbf{D}etection. ``Granularity'' denotes evaluation at the \textbf{R}esponse, \textbf{S}egment, or \textbf{C}laim level. ``Rationale'' indicates whether the benchmark provides explanatory justifications.}
    \label{tab:benchmark_cmp}
    \small
	\begin{adjustbox}{width=1.0\textwidth}
		\begin{tabular}{lcccccc}
			\toprule
			\textbf{Benchmark} & \textbf{Function} & \textbf{Granularity} & \textbf{\# Instances} & \textbf{Task} & \textbf{\# Mod.} & \textbf{Rationale}
			\\
			\midrule
			QAGS \citep{wang2020askingansweringquestionsevaluate} & C & R & 474 & T2T & 1 & \errormark \\
			HaluEval \citep{li2023haluevallargescalehallucinationevaluation} & D & R & 30,000 & T2T & 1 & \errormark \\
			POPE \citep{li2023evaluating} & D & R & 500 & I2T  & 2 & \errormark \\
			AMBER \citep{wang2024amberllmfreemultidimensionalbenchmark} & D & R & 1,004 & I2T & 2  & \errormark \\
            FactVC \citep{liu2023modelshallucinationsevaluatingfactuality} & D & R & 1,800 & V2T & 2 & \errormark \\
            AHLALM \citep{nishimura2024audio} & D & R & 1,000 & A2T & 2 & \errormark \\
			SoraDetector \citep{chu2024soradetectorunifiedhallucination} & D & R & 50 & T2V & 2 & \errormark \\
			MHaluBench \citep{chen2024unifiedhallucinationdetectionmultimodal} & D & R, S, C & 420 & T2I, I2T & 2 & \correctmark \\
			\midrule
			\textbf{OmniHallu-Bench (Ours)} & D & R, S, C & 10,000 & T2I, T2V, T2A, I2T, V2T, A2T & 4 & \correctmark \\
			\bottomrule
		\end{tabular}
	\end{adjustbox}
\end{table*}

\section{Related Work}

\paragraph{Hallucination in MLLMs.}
Hallucinations manifest across all MLLM modalities. In vision-language models, generated descriptions may mention objects absent from the image \citep{li2023evaluating}. Video-language models exhibit intrinsic and extrinsic hallucinations \citep{wang2024videohallucer,huang2026no}, while audio-video language models may ignore acoustic content and describe audio primarily from visual evidence \citep{nishimura2024audio}. In generation tasks, text-to-image models often fail on compositional prompt alignment \citep{Bakr2023HRSBenchHR,10847875}, and text-to-video models lack temporal coherence \citep{chu2024soradetectorunifiedhallucination,rawte2024vibe}. Despite this breadth of modality-specific work, most prior research has studied each modality and task type independently, limiting insight into cross-modal regularities.

\paragraph{Hallucination Detection and Evaluation.}
Detection methods have evolved from simple self-consistency checks \citep{manakul2023selfcheckgpt,miao2023selfcheckusingllmszeroshot} to structured multi-step pipelines. UNIHD \citep{chen2024unifiedhallucinationdetectionmultimodal} introduced claim decomposition and verification for image--text tasks, while FactVC \citep{liu2023modelshallucinationsevaluatingfactuality} proposed factuality metrics for video captioning. CrossCheckGPT \citep{sun2024crosscheckgpt} ranks systems through reference-free cross-system consistency across text, image, and audio-visual domains, and video-SALMONN~2 \citep{tang2025videosalmonn2} uses preference optimization to mitigate errors in audio-visual captioning. These methods target system ranking, claim-level evaluation, or hallucination mitigation, while benchmarks such as POPE \citep{li2023evaluating}, AMBER \citep{wang2024amberllmfreemultidimensionalbenchmark}, and MHaluBench \citep{chen2024unifiedhallucinationdetectionmultimodal} remain restricted to a single modality or modality pair. Our work studies claim-level detection across six bidirectional tasks; Table~\ref{tab:benchmark_cmp} summarizes the corresponding benchmark coverage.

\paragraph{Tool-Augmented and Multi-Agent LLM Systems.}
Toolformer \citep{schick2023toolformerlanguagemodelsteach} trains language models to decide which external tools to call and how to incorporate their outputs. Grounding DINO \citep{ren2024groundingdino15advance} enables zero-shot object verification, and DoraemonGPT \citep{yang2024doraemongpt} reformulates video understanding into tool invocations. Our framework extends this paradigm to hallucination detection across modalities with unified reasoning-based aggregation.

\section{Task Formulation and Hallucination Taxonomy}
\label{sec:prelim}

\paragraph{Unified Formulation.}
Let $\mathcal{T}$, $\mathcal{I}$, $\mathcal{V}$, $\mathcal{A}$ denote textual, image, video, and audio data. An MLLM maps input $\mathbf{x} \in \{\mathcal{T}, \mathcal{I}, \mathcal{V}, \mathcal{A}\}$ to output $\hat{y} \in \{\mathcal{T}, \mathcal{I}, \mathcal{V}, \mathcal{A}\}$. We address \emph{comprehension} tasks ($\mathbf{x} \in \{\mathcal{I}, \mathcal{V}, \mathcal{A}\} \to \hat{y} \in \mathcal{T}$), and \emph{generation} tasks ($\mathbf{x} \in \mathcal{T} \to \hat{y} \in \{\mathcal{I}, \mathcal{V}, \mathcal{A}\}$). Comprehension hallucinations reside in the generated text, while generation hallucinations manifest as semantic discrepancies between the produced media and the prompt.

\paragraph{Definition.}
An output $\hat{y}$ is hallucinated if it contains any semantic claim that is unsupported by or contradicts the input $\mathbf{x}$. Formally, let $\mathcal{G}$ denote the set of ground-truth semantic elements derivable from $\mathbf{x}$:
\[
\Hallucinate(\hat{y}\mid \mathbf{x}) =
\begin{cases}
1 & \text{if } \exists\;\phi(\hat{y}) \notin \mathcal{G} \\
0 & \text{otherwise,}
\end{cases}
\]
where $\phi(\hat{y})$ denotes any semantic claim extractable from $\hat{y}$.

\paragraph{Hallucination Taxonomy.}
We define four types applicable across all modalities: (1) \textbf{Object}: non-existent entities introduced; (2) \textbf{Attribute}: misrepresented properties such as color, size, or timbre; (3) \textbf{Relation}: incorrectly stated spatial, temporal, or causal relationships; (4) \textbf{Event}: misrepresented event-level details, including temporal ordering errors or fabricated actions. Each type manifests across all six tasks, enabling direct cross-modal comparison (\S\ref{sec:exp}).

\section{OmniHallu-Bench}
\label{sec:bench}

\begin{figure*}[t]
    \centering
    \includegraphics[width=1.0\textwidth]{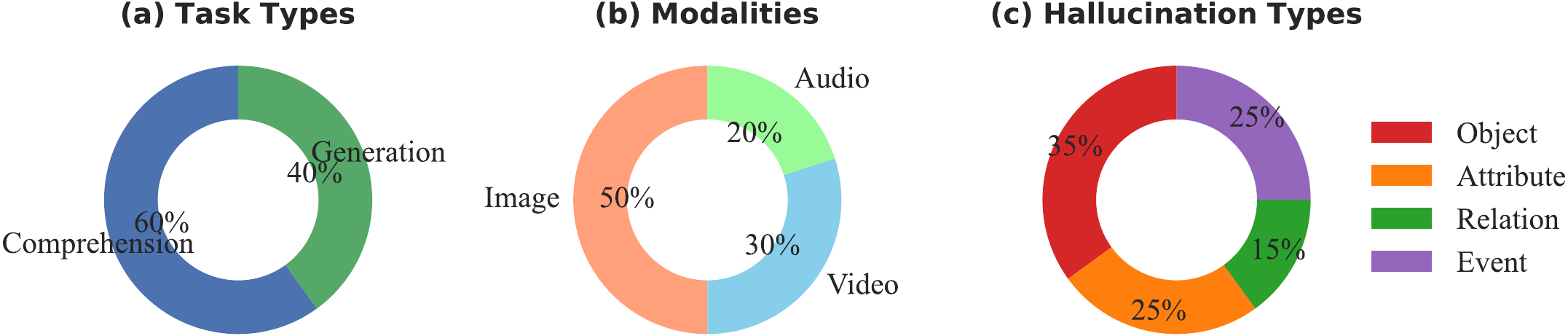}
    \caption{Statistics of OmniHallu-Bench: distribution across modalities, tasks, and hallucination types.}
    \label{fig:dataset_stats}
\end{figure*}

\paragraph{Design Principles.}
OmniHallu-Bench comprises 10,000 samples with stratified coverage across modalities and tasks. Comprehension tasks account for 60\% and generation tasks for 40\%. Image, video, and audio samples follow a 5:3:2 ratio. Hallucination types are distributed as: object (35\%), attribute (25\%), event (25\%), and relation (15\%).

\paragraph{Comprehension Tasks.}
For I2T, we draw from COCO Captions \citep{chen2015microsoftcococaptions}, Nocaps \citep{Agrawal_2019}, and Flickr30k \citep{plummer2016flickr30kentitiescollectingregiontophrase}, with model outputs generated by InternVL2.5-78B \citep{chen2024expanding}, Qwen2.5-VL-72B \citep{bai2025qwen25vl}, GPT-4.1 \citep{openai2025gpt41}, and Gemini-2.5-Pro \citep{comanici2025gemini}. For V2T, we sample from MSVD \citep{Chen_2022}, MSRVTT \citep{7780940}, and VATEX \citep{wang2020vatexlargescalehighqualitymultilingual}, using InternVL2.5-78B, Qwen2.5-VL-72B, VideoLLaMA3 \citep{zhang2025videollama3frontiermultimodal}, and LLaVA-OneVision \citep{li2024llavaonevisioneasyvisualtask}. For A2T, we use AudioCaps \citep{kim-etal-2019-audiocaps}, ClothoV2 \citep{drossos2019clothoaudiocaptioningdataset}, and AudioSetCaps \citep{bai2024audiosetcapsenrichedaudiocaptiondataset}, with outputs from Qwen2-Audio-7B-Instruct \citep{chu2024qwen2audio}, GAMA \citep{ghosh2024gamalargeaudiolanguagemodel}, Pengi \citep{deshmukh2024pengiaudiolanguagemodel}, and SALMONN \citep{tang2024salmonn}.

\paragraph{Generation Tasks.}
For T2I, prompts from T2I-CompBench++ \citep{10847875} and HRS-Bench \citep{Bakr2023HRSBenchHR} are used to generate images via DALL-E 3, Stable Diffusion 3.5 Large, and Midjourney v6 \citep{betker2023dalle3,stabilityai2024sd35,midjourney2023v6}. For T2V, prompts from T2V-CompBench \citep{sun2025t2vcompbenchcomprehensivebenchmarkcompositional} and FETV \citep{Liu2023FETVAB} drive generation via Open-Sora 1.2 \citep{zheng2024opensorademocratizingefficientvideo} and CogVideoX-5B \citep{yang2024cogvideoxtexttovideodiffusionmodels}. For T2A, prompts from WavText5K \citep{deshmukh2022audioretrievalwavtext5kclap}, FSD50K \citep{fonseca2022fsd50kopendatasethumanlabeled}, and SoundDescs \citep{Koepke_2023} generate audio via Make-an-Audio \citep{huang2023makeanaudiotexttoaudiogenerationpromptenhanced}, AudioGPT \citep{huang2023audiogptunderstandinggeneratingspeech}, and AudioLCM \citep{liu2024audiolcmtexttoaudiogenerationlatent}. Multiple generators are used for each task to reduce dependence on model-specific artifacts.

\paragraph{Annotation and Quality Control.}
Samples undergo a structured atomic claim decomposition using Chain-of-Thought prompting \citep{cot} with self-reflection verification. Three trained annotators independently review each sample; a sample is retained only upon full consensus. Annotators first validate the decomposed claims for semantic fidelity, then classify each as hallucinatory or non-hallucinatory. Approximately 24.3\% of initial samples were removed due to disagreement. Before consensus filtering, Fleiss' $\kappa$ is 0.89 (image), 0.86 (video), and 0.83 (audio). The final dataset contains 4,000 human-curated and 6,000 model-generated (human-audited) samples.

\paragraph{Dataset Statistics.}
The average number of atomic claims per sample is 4.2 (I2T), 3.8 (T2I), 5.6 (V2T), 4.1 (T2V), 3.5 (A2T), and 3.1 (T2A). The overall hallucination rate is 42.8\%, compared with 48.1\% for generation tasks and 39.2\% for comprehension tasks.

\section{Multi-Agent Hallucination Detection}
\label{sec:method}

\begin{figure*}[t]
    \centering
    \includegraphics[width=1.0\textwidth]{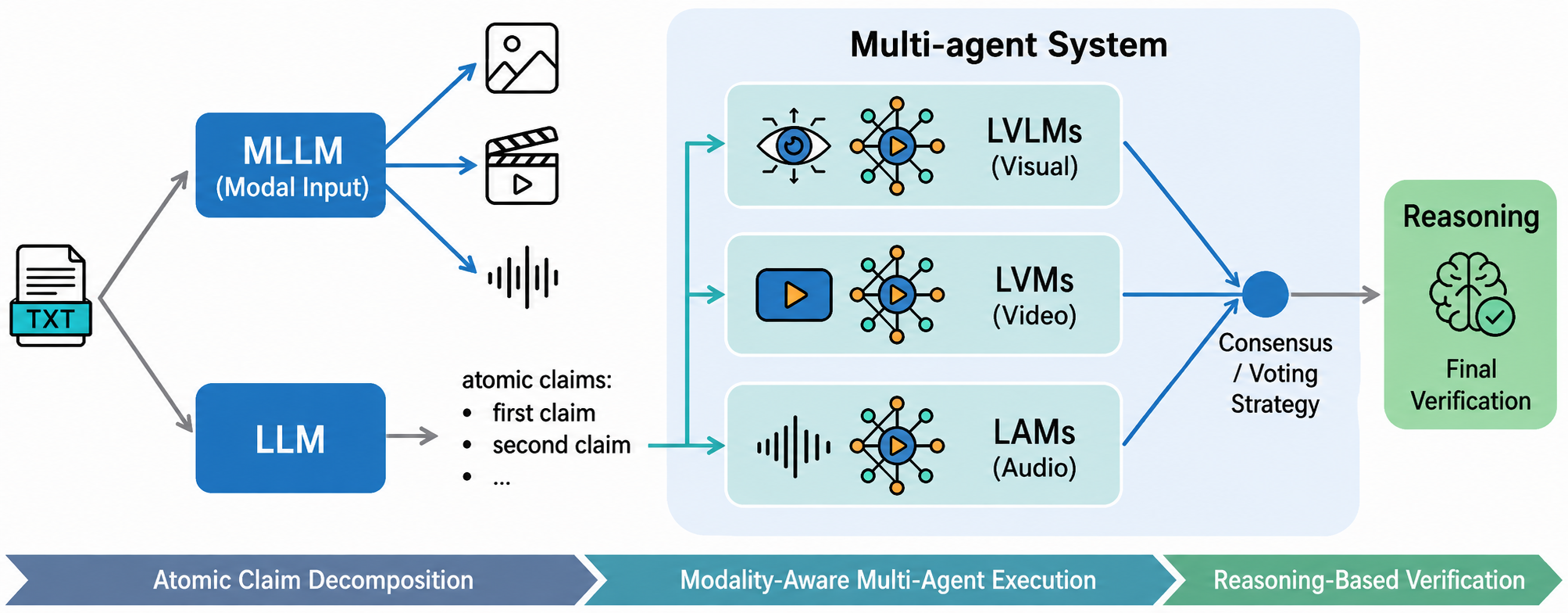}
    \caption{Overview of the OmniHallu multi-agent framework. Model outputs are decomposed into atomic claims, verified by modality-specific experts, and aggregated through reasoning-based decision making.}
    \label{fig:framework}
\end{figure*}

Our framework consists of three stages: atomic claim decomposition, modality-aware expert verification, and reasoning-based aggregation (Figure~\ref{fig:framework}).

\paragraph{Atomic Claim Decomposition (ACD).}
We decompose the target text (model-generated captions for comprehension, input prompts for generation) into atomic claims using GPT-4.1. Each sample $(y, \{c_1, \ldots, c_{n_y}\})$ consists of text $y$ and corresponding claims, where each $c_i$ is a semantically discrete, grammatically self-contained, verifiable statement. For comprehension tasks, $y$ is the model-generated caption; for generation tasks, $y$ is the input prompt whose claims must be verified against the generated media. We compare GPT-4.1 claims against human reference claims on a 300-sample subset (Table~\ref{tab:decomp_quality}); mean coverage is 0.92 and mean redundancy is 1.15.

\begin{table}[t]
\centering
\caption{Atomic claim decomposition quality against human references, measured by coverage, redundancy, and agreement.}
\label{tab:decomp_quality}
\small
\setlength{\tabcolsep}{8pt}
\begin{tabular}{lccc}
\toprule
\textbf{Modality} & \textbf{Cov. $\uparrow$} & \textbf{Red. $\downarrow$} & \textbf{Agr. $\uparrow$} \\
\midrule
Image (I2T/T2I) & 0.96 & 1.08 & 0.92 \\
Video (V2T/T2V) & 0.92 & 1.15 & 0.88 \\
Audio (A2T/T2A) & 0.89 & 1.21 & 0.85 \\
\midrule
\textbf{Average} & \textbf{0.92} & \textbf{1.15} & \textbf{0.88} \\
\bottomrule
\end{tabular}
\end{table}

\paragraph{Modality-Aware Expert Verification.}
Different modalities and hallucination types require specialized verification:

\textit{Image tasks (I2T, T2I).} Object hallucinations are verified using Grounding DINO 1.5 Pro \citep{ren2024groundingdino15advance} for open-set detection. Attribute, relation, and event hallucinations are assessed by an ensemble of MLLMs (Qwen2.5-VL-72B, InternVL2.5-78B, GPT-4.1), where each model independently evaluates the claim against visual evidence. For T2I tasks, the text prompt serves as ground truth and experts verify whether the generated image faithfully reflects each prompted claim.

\textit{Video tasks (V2T, T2V).} Following DoraemonGPT \citep{yang2024doraemongpt}, each atomic claim is reformulated into a targeted QA query via GPT-4.1, enabling temporal decomposition and frame-level evidence extraction. The default three-expert configuration uses Qwen2.5-VL-72B, InternVL2.5-78B, and VideoLLaMA3 \citep{zhang2025videollama3frontiermultimodal}, which can attend to specific temporal segments.

\textit{Audio tasks (A2T, T2A).} The default three-expert configuration uses Qwen2-Audio-7B-Instruct, GAMA \citep{ghosh2024gamalargeaudiolanguagemodel}, and SALMONN \citep{tang2024salmonn}.

Within each modality, expert judgments are summarized by equal-weight majority voting, chosen for transparency and because no reliable cross-model confidence calibration metric is available. The vote summary and the individual evidence traces are then passed to the final reasoning-based aggregator described below (see Appendix~\ref{app:voting}).

\paragraph{Reasoning-Based Aggregation.}
Expert verification results and atomic claims are consolidated by GPT-5.2 in the main experiments; the controlled comparison in Table~\ref{tab:reasoning_model} replaces it with GPT-4.1 \citep{openai2025gpt41,openai2025gpt52}. The aggregator receives each expert's judgment, supporting evidence, and confidence signal, and outputs a final label with an explanatory rationale.

\paragraph{Trainable Verifier via Preference Optimization.}
\label{sec:verifier}
To reduce dependence on expensive expert ensembles, we train a compact claim-level verifier $\pi_{\psi}$ (initialized from Qwen2.5-VL-7B) that outputs a judgment conditioned on the task input and an atomic claim:
\begin{equation}
\label{eq:verifier}
\pi_{\psi}(\ell \mid \mathbf{x}, c_i),\;\; \ell \in \{\texttt{SUP},\texttt{UNSUP},\texttt{ABS}\}.
\end{equation}
\texttt{UNSUP} maps to hallucinated, \texttt{SUP} to non-hallucinated, and \texttt{ABS} (abstain) triggers full expert verification.

We define a reward $R = \lambda_{\text{lab}} R_{\text{lab}} + \lambda_{\text{ev}} R_{\text{ev}} + \lambda_{\text{cal}} R_{\text{cal}}$, using $(\lambda_{\text{lab}},\lambda_{\text{ev}},\lambda_{\text{cal}})=(1.0,0.7,0.3)$. Here, $R_{\text{lab}}$ rewards ground-truth match, $R_{\text{ev}}$ rewards consistency with multi-agent consensus, and $R_{\text{cal}}$ penalizes overconfidence when expert signals conflict. We train via Group Relative Policy Optimization (GRPO) \citep{shao2024deepseekmath}:
\begin{equation}
\label{eq:grpo}
\mathcal{L}_{\text{GRPO}} = -\mathbb{E}\Bigg[\frac{1}{K}\sum_{k=1}^{K} \hat{A}_k\,\log \pi_{\psi}(y_k\mid \mathbf{x},c_i)\Bigg],
\end{equation}
where $\hat{A}_k = (R_k-\mu(R))/(\sigma(R)+\epsilon)$ is the normalized advantage over $K$ sampled judgments. GRPO uses group-level relative advantages instead of explicit preference pairs. The trained verifier integrates into the pipeline as a low-cost filter: confident predictions (probability $>0.85$) skip expert calls, while uncertain samples (\texttt{ABS}) fall through to the full ensemble.

\section{Experiments}
\label{sec:exp}

\paragraph{Setup.}
We follow the evaluation protocol of UNIHD \citep{chen2024unifiedhallucinationdetectionmultimodal}, computing precision (P), recall (R), and F1 for both hallucinatory and non-hallucinatory categories at claim level, along with accuracy and macro-averaged F1 (Mac.F1). Mac.F1 is our primary metric as it balances detection of both hallucinatory and non-hallucinatory claims. Baselines include: (1) \textit{Self-Check} \citep{miao2023selfcheckusingllmszeroshot}, which uses a single MLLM's chain-of-thought self-verification without external tools; and (2) \textit{UNIHD} \citep{chen2024unifiedhallucinationdetectionmultimodal}, a multi-step pipeline applicable only to image tasks. For video and audio tasks where UNIHD is inapplicable, we compare against Self-Check with the strongest available MLLMs per modality. Our full framework uses GPT-5.2 \citep{openai2025gpt52} as the reasoning model; we additionally report GPT-4.1 results (Table~\ref{tab:reasoning_model}) to enable direct comparison at equal model capacity.
For the verifier, we use a disjoint 6,000/1,000/3,000-sample train/dev/test split with no source-media overlap. The trainable verifier is initialized from Qwen2.5-VL-7B and trained for 3 epochs with learning rate $1\times10^{-5}$ and GRPO group size $K=8$.

\begin{table*}[t]
\caption{Hallucination detection results across six tasks. P/R/F1 are reported for hallucinatory (H) and non-hallucinatory (NH) categories; Mac.F1 is the macro-average. All values are claim-level percentages. Our framework uses GPT-5.2 as the reasoning model; baselines use their respective models as noted. Best results in \textbf{bold}.}
\label{tab:results_all}
\setlength{\tabcolsep}{2.0mm}
\centering
\resizebox{1.0\textwidth}{!}{
\begin{tabular}{@{}llccccccccccc@{}}
\toprule
\multicolumn{2}{c}{\multirow{2}{*}{\bfseries Method}} & \multicolumn{3}{c}{\bfseries Hallucinatory} & \multicolumn{3}{c}{\bfseries Non-Hallucinatory} & \multicolumn{4}{c}{\bfseries Overall} \\
\cmidrule(r){3-5} \cmidrule(r){6-8} \cmidrule(r){9-12}
& & P & R & F1 & P & R & F1 & Acc. & P & R & Mac.F1 \\
\midrule
\multicolumn{12}{c}{\cellcolor{gray!10}\bfseries Image-to-Text (I2T)} \\
\midrule
\multirow{2}{*}{Gemini-2.5-Pro}
& Self-Check & 82.75 & 64.12 & 72.20 & 66.12 & 81.31 & 72.90 & 72.48 & 74.44 & 72.72 & 72.55 \\
& UNIHD & 83.94 & 68.21 & 75.27 & 69.92 & 81.45 & 75.24 & 75.92 & 76.93 & 74.83 & 75.26 \\
\multirow{2}{*}{GPT-4.1}
& Self-Check & 79.32 & 73.92 & 76.52 & 74.31 & 80.54 & 77.30 & 76.21 & 76.82 & 77.23 & 76.91 \\
& UNIHD & 81.02 & 77.45 & 79.20 & 77.23 & 79.92 & 78.55 & 78.12 & 79.13 & 78.69 & 78.88 \\
\bfseries Ours & & \bf 84.65 & \bf 81.34 & \bf 82.96 & \bf 83.15 & \bf 82.72 & \bf 82.93 & \bf 82.58 & \bf 83.90 & \bf 82.03 & \bf 82.95 \\
\midrule
\multicolumn{12}{c}{\cellcolor{gray!10}\bfseries Text-to-Image (T2I)} \\
\midrule
\multirow{2}{*}{Gemini-2.5-Pro}
& Self-Check & 81.12 & 59.82 & 68.86 & 63.41 & 78.74 & 70.23 & 70.98 & 72.27 & 69.28 & 69.55 \\
& UNIHD & 82.64 & 64.52 & 72.48 & 65.92 & 80.32 & 72.41 & 73.84 & 74.28 & 72.42 & 72.45 \\
\multirow{2}{*}{GPT-4.1}
& Self-Check & 78.22 & 71.94 & 74.95 & 72.63 & 78.32 & 75.37 & 74.52 & 75.43 & 75.13 & 75.16 \\
& UNIHD & 79.88 & 76.54 & 78.18 & 78.65 & 78.80 & 78.72 & 78.64 & 79.27 & 77.67 & 78.45 \\
\bfseries Ours & & \bf 83.21 & \bf 80.92 & \bf 82.04 & \bf 81.74 & \bf 81.42 & \bf 81.58 & \bf 81.40 & \bf 82.48 & \bf 81.17 & \bf 81.81 \\
\midrule
\multicolumn{12}{c}{\cellcolor{gray!10}\bfseries Video-to-Text (V2T)} \\
\midrule
InternVL2.5-78B & Self-Check & 74.12 & 60.82 & 66.82 & 65.34 & 69.24 & 67.24 & 65.46 & 69.73 & 65.03 & 67.03 \\
Qwen2.5-VL-72B & Self-Check & 76.58 & 65.12 & 70.40 & 68.42 & 71.32 & 69.84 & 68.36 & 72.50 & 68.22 & 70.12 \\
\bfseries Ours & & \bf 78.92 & \bf 75.42 & \bf 77.13 & \bf 77.38 & \bf 76.92 & \bf 77.15 & \bf 77.02 & \bf 78.15 & \bf 76.17 & \bf 77.14 \\
\midrule
\multicolumn{12}{c}{\cellcolor{gray!10}\bfseries Text-to-Video (T2V)} \\
\midrule
InternVL2.5-78B & Self-Check & 72.94 & 58.12 & 64.71 & 62.72 & 68.34 & 65.41 & 63.48 & 67.83 & 63.23 & 65.06 \\
Qwen2.5-VL-72B & Self-Check & 74.75 & 63.42 & 68.63 & 67.10 & 69.42 & 68.24 & 66.58 & 70.93 & 66.42 & 68.44 \\
\bfseries Ours & & \bf 77.62 & \bf 74.12 & \bf 75.83 & \bf 75.92 & \bf 74.92 & \bf 75.42 & \bf 75.22 & \bf 76.77 & \bf 74.52 & \bf 75.63 \\
\midrule
\multicolumn{12}{c}{\cellcolor{gray!10}\bfseries Audio-to-Text (A2T)} \\
\midrule
GAMA  & Self-Check & 71.34 & 56.72 & 63.19 & 62.94 & 68.02 & 65.38 & 62.54 & 67.14 & 62.37 & 64.29 \\
Qwen2-Audio-7B  & Self-Check & 73.48 & 59.24 & 65.57 & 64.82 & 69.28 & 66.98 & 64.42 & 69.15 & 64.26 & 66.28 \\
\bfseries Ours & & \bf 76.38 & \bf 72.12 & \bf 74.19 & \bf 74.64 & \bf 73.42 & \bf 74.02 & \bf 73.82 & \bf 75.51 & \bf 72.77 & \bf 74.11 \\
\midrule
\multicolumn{12}{c}{\cellcolor{gray!10}\bfseries Text-to-Audio (T2A)} \\
\midrule
GAMA & Self-Check & 70.15 & 55.48 & 61.95 & 61.72 & 67.42 & 64.44 & 61.64 & 65.94 & 61.45 & 63.20 \\
Qwen2-Audio-7B & Self-Check & 72.48 & 58.34 & 64.62 & 63.92 & 68.36 & 66.06 & 63.54 & 68.20 & 63.35 & 65.34 \\
\bfseries Ours & & \bf 75.48 & \bf 71.52 & \bf 73.45 & \bf 73.74 & \bf 72.92 & \bf 73.33 & \bf 73.12 & \bf 74.61 & \bf 72.22 & \bf 73.39 \\
\bottomrule
\end{tabular}
}
\end{table*}

\subsection{Main Results}

Table~\ref{tab:results_all} presents claim-level results across all six tasks. Our multi-agent framework consistently outperforms all baselines, with Mac.F1 improvements ranging from +3.4 to +8.1 points over the strongest baseline per task. The improvements are statistically significant across all tasks ($p<0.01$, bootstrap test with 10,000 resamples). We discuss the results along three dimensions: modality, task type, and hallucination category.

\textbf{Image tasks} exhibit the highest performance overall (Mac.F1: 81.8--83.0). With the UNIHD pipeline available, baselines are relatively strong, yet our framework still achieves +4.1 (I2T) and +3.4 (T2I) points over GPT-4.1 UNIHD.

\textbf{Video tasks} show moderate performance (Mac.F1: 75.6--77.1), reflecting the added complexity of temporal reasoning. Our framework achieves +7.0 (V2T) and +7.2 (T2V) points over the best Self-Check baseline, demonstrating the value of multi-agent verification for temporal content.

\textbf{Audio tasks} are the most challenging (Mac.F1: 73.4--74.1), consistent with limited expert tool availability for auditory understanding. Our framework still provides +7.8 (A2T) and +8.1 (T2A) points improvement, the largest absolute gains across all tasks, demonstrating that the multi-agent architecture is particularly valuable when individual models are weaker.

Within each modality, comprehension tasks consistently outperform their generation counterparts (e.g., I2T 82.95 vs.\ T2I 81.81; V2T 77.14 vs.\ T2V 75.63), as comprehension verification can directly ground claims against the source media, whereas generation verification must assess whether the produced media faithfully reflects the textual prompt. This modality-dependent performance gradient (image $>$ video $>$ audio) results from two interacting factors: the intrinsic complexity of each modality and the varying maturity of available expert models, where vision experts (70B+ parameters) far exceed current audio experts.

\subsection{Ablation Studies}

\begin{table}[t]
    \centering
    \caption{Module ablation results in Mac.F1. ACD and MV denote Atomic Claim Decomposition and Multi-expert Voting, respectively.}
    \label{tab:ablation}
    \small
    \setlength{\tabcolsep}{3.5pt}
    \renewcommand{\arraystretch}{0.98}
    \begin{tabular}{lccc}
        \toprule
        \textbf{Task} & \textbf{Full} & \textbf{w/o ACD} & \textbf{w/o MV} \\
        \midrule
        I2T & 82.95 & 75.02$_{-7.93}$ & 77.64$_{-5.31}$ \\
        T2I & 81.81 & 74.31$_{-7.50}$ & 76.55$_{-5.26}$ \\
        V2T & 77.14 & 70.10$_{-7.04}$ & 72.34$_{-4.80}$ \\
        T2V & 75.63 & 68.42$_{-7.21}$ & 70.71$_{-4.92}$ \\
        A2T & 74.11 & 67.05$_{-7.06}$ & 69.24$_{-4.87}$ \\
        T2A & 73.39 & 66.81$_{-6.58}$ & 68.92$_{-4.47}$ \\
        \bottomrule
    \end{tabular}
\end{table}

\paragraph{Module Ablation.}
Table~\ref{tab:ablation} shows that removing ACD causes the largest degradation (-6.6 to -7.9 points), confirming the importance of structured claim decomposition for precise hallucination localization. Without ACD, the reasoning model must assess hallucination at the response level, losing fine-grained pinpointing. Removing multi-expert voting (MV) reduces performance by 4.5--5.3 points uniformly across modalities, validating the ensemble verification strategy. The consistent gap between ACD and MV degradations across all six tasks indicates that precise claim-level granularity contributes more to detection accuracy than ensemble diversity, underscoring that decomposition quality is the primary bottleneck in the pipeline.

\begin{table}[t]
\centering
\caption{Reward ablation for the GRPO-trained verifier.}
\label{tab:reward_ablation}
\small
\setlength{\tabcolsep}{8pt}
\begin{tabular}{lcc}
\toprule
\textbf{Setting} & \textbf{Mac.F1 $\uparrow$} & \textbf{ECE $\downarrow$} \\
\midrule
Full Reward $R$ & \textbf{78.94} & \textbf{0.042} \\
w/o $R_{\text{lab}}$ & 74.21 & 0.085 \\
w/o $R_{\text{ev}}$ & 71.55 & 0.112 \\
w/o $R_{\text{cal}}$ & 77.10 & 0.154 \\
\bottomrule
\end{tabular}
\end{table}

\paragraph{Reward Ablation.}
Table~\ref{tab:reward_ablation} shows that removing $R_{\text{ev}}$ (expert evidence consistency) causes the largest Mac.F1 drop (-7.4), underscoring the importance of aligning the verifier with multi-agent consensus. Removing $R_{\text{cal}}$ (calibration penalty) yields the largest ECE increase (0.042$\to$0.154), indicating that the verifier becomes overconfident without calibration pressure. All three components serve complementary roles: $R_{\text{lab}}$ removal yields a moderate decline (-4.7 Mac.F1) with ECE rising to 0.085, confirming that ground-truth supervision contributes to both accuracy and calibration, though its calibration effect is secondary to the dedicated $R_{\text{cal}}$ penalty.

\subsection{Reasoning Model and Cost Analysis}

\begin{table}[t]
\centering
\caption{Impact of reasoning model choice on Mac.F1.}
\label{tab:reasoning_model}
\small
\setlength{\tabcolsep}{8pt}
\begin{tabular}{lcc}
\toprule
\textbf{Task} & \textbf{GPT-5.2} & \textbf{GPT-4.1} \\
\midrule
I2T & 82.95 & 78.44 \\
T2I & 81.81 & 76.79 \\
V2T & 77.14 & 71.83 \\
T2V & 75.63 & 69.91 \\
A2T & 74.11 & 67.56 \\
T2A & 73.39 & 67.23 \\
\bottomrule
\end{tabular}
\end{table}

\paragraph{Reasoning Model.}
Table~\ref{tab:reasoning_model} shows that GPT-5.2 consistently outperforms GPT-4.1 as the reasoning backbone by 4.5--6.6 points, with the gap more pronounced on video and audio tasks where evidence synthesis across temporal and acoustic dimensions is harder. We acknowledge that the main results (Table~\ref{tab:results_all}) use GPT-5.2, giving our framework a stronger backbone than baselines. Even with GPT-4.1, however, our framework outperforms baselines on video and audio tasks by 1.3--1.9 points, demonstrating that the multi-agent architecture provides genuine value beyond model strength alone. Notably, the gap widens from 4.5--5.0 points on image tasks to 6.2--6.6 points on audio tasks, suggesting that stronger reasoning models compensate more effectively when expert tools provide weaker or more ambiguous perceptual evidence.

\begin{table}[t]
\centering
\caption{Computational cost analysis per sample, averaged across tasks.}
\label{tab:cost}
\small
\setlength{\tabcolsep}{3pt}
\begin{tabular}{lccc}
\toprule
\textbf{Method} & \textbf{API Calls} & \textbf{Time (s)} & \textbf{Est. Cost} \\
\midrule
Self-Check (GPT-4.1) & 1 & $\sim$3 & $\sim$\$0.02 \\
UNIHD (Image only) & 4--6 & $\sim$12 & $\sim$\$0.08 \\
Ours (Full pipeline) & 5--8 & $\sim$15 & $\sim$\$0.12 \\
Ours + Filter ($\pi_\psi$) & 1.7--2.8 & $\sim$6 & $\sim$\$0.05 \\
\bottomrule
\end{tabular}
\end{table}

\paragraph{Computational Cost.}
Table~\ref{tab:cost} reports computational overhead. The full pipeline requires 5--8 API calls at $\sim$\$0.12 per sample. The verifier filter reduces API calls to 1.7--2.8 (66\% reduction) and cost to $\sim$\$0.05 per sample. For the full OmniHallu-Bench, the estimated total cost is $\sim$\$1,200 for the full pipeline or $\sim$\$500 with the verifier filter, offering a practical trade-off for large-scale deployment. The latency reduction from $\sim$15s to $\sim$6s per sample further enables near-real-time hallucination feedback in interactive applications.

\subsection{Trainable Verifier Integration}

\begin{table}[t]
\centering
\caption{Trainable verifier integration modes. ``Expert Calls'' indicates the fraction of samples requiring full ensemble invocation.}
\label{tab:verifier}
\small
\setlength{\tabcolsep}{4pt}
\begin{tabular}{lccc}
\toprule
\textbf{Setting} & \textbf{Training} & \textbf{Mac.F1} & \textbf{Expert$\downarrow$} \\
\midrule
Verifier-only & SFT  & 71.42 & 0\% \\
Verifier-only & DPO              & 76.85 & 0\% \\
Verifier-only & GRPO             & 78.94 & 0\% \\
Filter (Hybrid) & GRPO           & 80.52 & 34.2\% \\
Aggregator      & GRPO           & \textbf{83.15} & 100\% \\
\bottomrule
\end{tabular}
\end{table}

Table~\ref{tab:verifier} compares integration strategies. The GRPO-trained verifier alone approaches the full pipeline without any expert calls, providing a practical option when latency or cost precludes expert invocation. The filter mode achieves 80.52 Mac.F1 while requiring expert calls for only 34.2\% of samples: the verifier handles easy cases (clear hallucinations or clearly supported claims) and defers ambiguous ones to the full ensemble. The aggregator mode uses $\pi_\psi$ to re-weight expert votes rather than replace them, achieving the highest performance (83.15) by improving robustness under expert disagreement. GRPO training consistently outperforms SFT (+7.5) and DPO (+2.1), validating the advantage of group-relative optimization for calibrated hallucination judgment. The three integration modes form a practical cost-performance spectrum, enabling practitioners to select an operating point based on deployment constraints: the standalone verifier for latency-sensitive applications, the filter mode for balanced cost-accuracy trade-offs, and the aggregator for high-stakes scenarios requiring maximum reliability.

\subsection{Fine-Grained Analysis}

\begin{figure}[t]
    \centering
    \includegraphics[width=1.0\columnwidth]{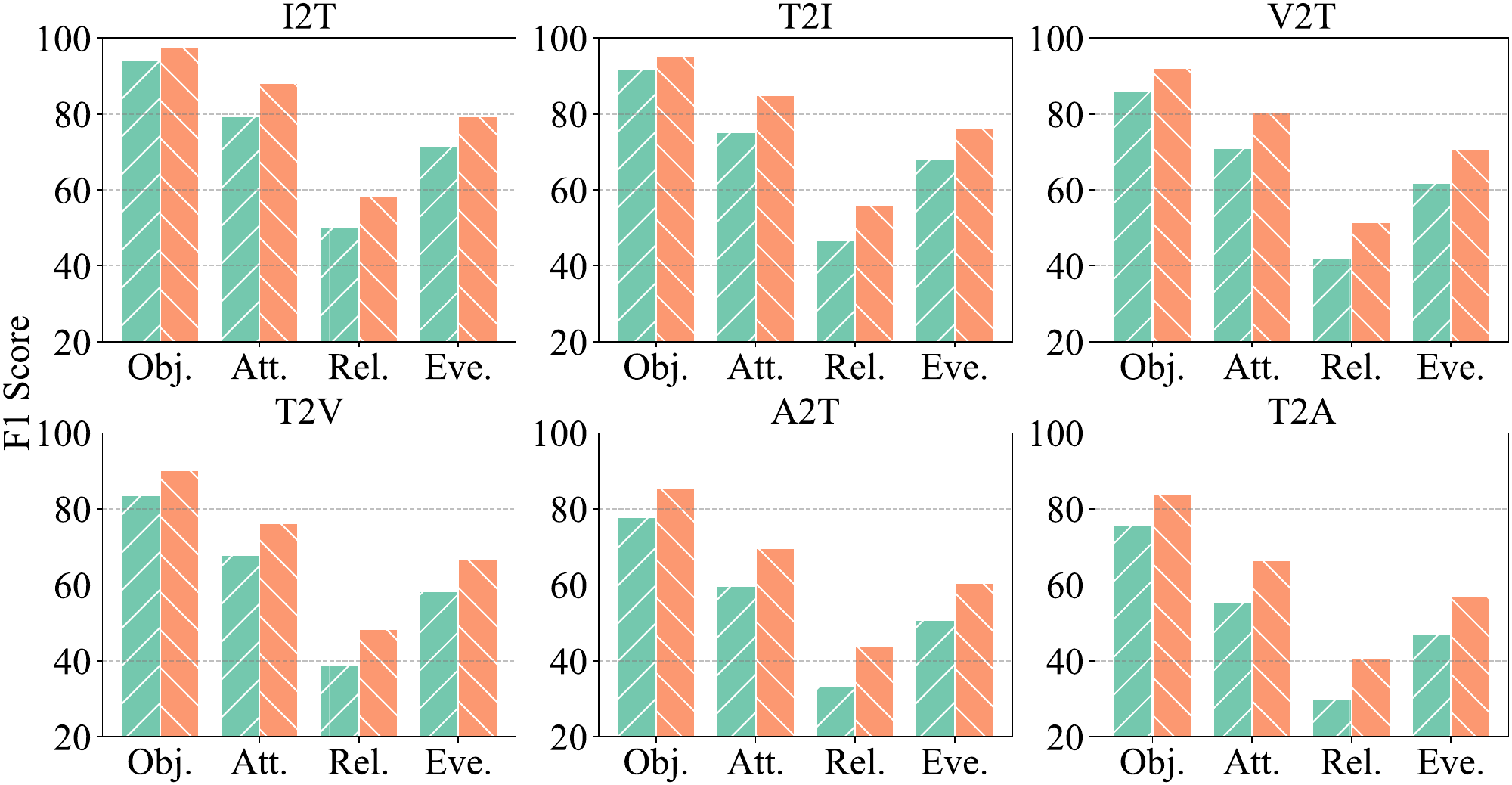}
\caption{Detection performance by hallucination type. Object hallucinations are easiest to detect; relation hallucinations are the most challenging across all modalities.}
    \label{fig:type_comparison}
\end{figure}

\paragraph{Hallucination Type Analysis.}
Figure~\ref{fig:type_comparison} reveals a consistent difficulty hierarchy across modalities on OmniHallu-Bench: object $>$ attribute $>$ event $>$ relation. Object hallucinations benefit from direct grounding tools, while attribute hallucinations require finer-grained perception. Event hallucinations demand temporal awareness, and relation hallucinations require compositional reasoning over multiple entities and their interactions, representing the most demanding verification task. This consistent ordering suggests that hallucination difficulty is primarily governed by the compositional complexity of claims rather than by modality-specific perceptual challenges alone. Our framework shows the largest improvements on relation hallucinations (+9.2 points on average), where multi-expert aggregation is most beneficial as individual experts often capture complementary relational evidence.

\paragraph{Cross-Modal Failure Patterns.}
Manual error analysis on 50 misclassified samples per task reveals distinct failure modes. \textit{Image}: small or heavily occluded objects and subtle attribute differences account for the majority of errors; object detectors frequently miss small-scale instances. \textit{Video}: temporal misalignment, including event ordering errors, omitted intermediate actions, and incorrect causal attributions, is the primary failure mode, particularly when evidence is distributed across non-adjacent frames requiring long-range temporal reasoning. \textit{Audio}: weak source separation and ambiguous acoustic cues are the main error sources; the ensemble voting strategy is vulnerable when all models share similar perceptual limitations. These patterns highlight modality-specific bottlenecks: image detection is primarily perception-limited, video detection is reasoning-limited, and audio detection is tool-limited. While image and video failures share perception-related origins rooted in visual understanding, audio failures arise from fundamentally different acoustic modeling limitations, suggesting that future improvements in each modality may require targeted architectural innovations. Addressing these modality-specific bottlenecks represents the most promising avenue for advancing cross-modal hallucination detection.

\section{Conclusion}
We presented OmniHallu and OmniHallu-Bench as a unified framework and claim-level benchmark for hallucination detection across cross-modal comprehension and generation. By placing diverse tasks and modalities under a common evaluation protocol, our work enables systematic comparison beyond modality-specific settings.
Our results show that claim decomposition, modality-aware evidence, and structured reasoning improve hallucination detection across heterogeneous scenarios. They also reveal persistent weaknesses in temporal and audio-grounded verification, and a clear difficulty shift from object and attribute errors to event and relation errors. These findings suggest that future progress will require stronger temporal, auditory, and compositional verification capabilities rather than generic detectors alone.

\section*{Limitations}

\paragraph{Taxonomy granularity and scope.}
Our four-type taxonomy provides a common cross-modal foundation but compresses modality-specific phenomena and compound errors. For example, video events could be divided into temporal ordering, action omission, and causal errors, while audio errors could distinguish source separation from timbre. Our text-centric formulation also excludes non-text pairings such as image-to-audio and video-to-image, which may require different decomposition strategies.

\paragraph{Model and expert dependence.}
Performance depends on the reasoning model (Table~\ref{tab:reasoning_model}) and the available experts. Vision experts are substantially larger than current audio experts, so cross-modal differences partly reflect tool capability rather than modality alone. Majority voting also cannot correct errors shared by all experts.

\section*{Ethical Considerations}
All benchmark sources are publicly available, and model-generated content is disclosed and human-audited. We recommend transparent documentation and human review before high-stakes use.

\bibliography{custom}

\appendix

\section{Voting Mechanism Design}
\label{app:voting}
We use equal-weight voting because the selected experts (e.g., GPT-4.1, Qwen2.5-VL, InternVL2.5) have comparable performance in their respective domains. Unequal weighting would require reliable cross-model calibration, whereas equal voting is transparent and requires no task-specific tuning.

\section{Additional Ablation Studies}
\label{app:ablation}

\begin{table}[ht]
\centering
\caption{Ablation on the number of expert agents (Mac.F1).}
\label{tab:num_agents}
\small
\setlength{\tabcolsep}{4pt}
\begin{tabular}{lccc}
\toprule
\textbf{Task} & \textbf{1 Expert} & \textbf{2 Experts} & \textbf{3 (Ours)} \\
\midrule
I2T   & 77.62 & 80.54 & 82.95 \\
T2I   & 76.74 & 78.92 & 81.81 \\
V2T   & 71.24 & 74.52 & 77.14 \\
T2V   & 69.85 & 72.96 & 75.63 \\
A2T   & 68.20 & 71.18 & 74.11 \\
T2A   & 67.54 & 70.02 & 73.39 \\
\bottomrule
\end{tabular}
\end{table}

\paragraph{Model Capacity.}
Replacing large experts (72B/78B) with small counterparts (7B/8B) on video tasks reduces Mac.F1 from 76.58/74.78 to 62.54/59.25.

\paragraph{Content Complexity.}
On T2V, hallucination prevalence increases with prompt complexity: 23.2\% (low density, $<$10s, 1 action), 28.7\% (medium, 10--20s, 2--3 actions), and 40.3\% (high, $>$20s, $\geq$4 actions).

\end{document}